\documentclass[runningheads]{llncs}
\usepackage{graphicx}
\usepackage{amsmath}
\usepackage{float}
\usepackage{algorithm2e}
\usepackage{cite}
\usepackage{hyperref}

\begin{document}
\title{Black Box Optimization Using QUBO and the Cross Entropy Method}
%
%
\author{Jonas Nüßlein\inst{1}\orcidID{0000-0001-7129-1237} \and
Christoph Roch\inst{1} \and
Thomas Gabor\inst{1} \and
Jonas Stein\inst{1} \and
Claudia Linnhoff-Popien\inst{1} \and
Sebastian Feld\inst{2}
}
\authorrunning{Nüßlein et al.}
%
\institute{Institute of Computer Science, LMU Munich, Germany \email{jonas.nuesslein@ifi.lmu.de} \and
Faculty of Electrical Engineering, Mathematics and Computer Science, TU Delft, Netherlands
}
\maketitle              
\begin{abstract}
Black-box optimization (BBO) can be used to optimize functions whose analytic form is unknown. A common approach to realising BBO is to learn a surrogate model which approximates the target black-box function which can then be solved via white-box optimization methods. In this paper, we present our approach \textit{BOX-QUBO}, where the surrogate model is a QUBO matrix. However, unlike in previous state-of-the-art approaches, this matrix is not trained entirely by regression, but mostly by classification between ``good'' and ``bad'' solutions. This better accounts for the low capacity of the QUBO matrix, resulting in significantly better solutions overall. We tested our approach against the state-of-the-art on four domains and in all of them \textit{BOX-QUBO} showed better results. A second contribution of this paper is the idea to also solve white-box problems, i.e. problems which could be directly formulated as QUBO, by means of black-box optimization in order to reduce the size of the QUBOs to the information-theoretic minimum. Experiments show that this significantly improves the results for MAX-$k$-SAT.

\end{abstract}
\section{Introduction}

The goal of black-box optimization is to minimize a function $E(x)$, where this function is not known analytically. Like an oracle, this function can only be queried for a given $x$: $y=E(x)$. A commonly used solution for this problem is to create a surrogate model $E'(x)$. This surrogate model is trained to provide the same outputs as $E(x)$. Since, unlike $E(x)$, $E'(x)$ is a white-box model, it is easier to search for the best solution $x^*=argmin_x\ E'(x)$. Black-box optimization then iterates between searching the best solution $x^*$ for the surrogate model $E'(x)$, asking the oracle for the actual value $y=E(x^*)$ and re-training the surrogate model $E'$ to output $y$ for the input $x^*$ using the mean squared error as the loss function: $loss = (E'(x^*) - y)^2$. We use the terms ``oracle'' and ``black-box function'' interchangeably in this paper.
\ \\
\ \\
Following the tradition of machine learning, we use the term `capacity' to refer to the amount of information a model can memorize. In black-box optimization, there is a trade-off between the capacity of the surrogate model (the higher the capacity of surrogate model $E'(x)$, the better it can approximate the actual values $y$) and the difficulty of the optimization (the higher the capacity of surrogate model $E'(x)$, the more difficult the optimization $x^*=argmin_x\ E'(x)$). A frequent choice for the model $E'(x)$ recently fell \cite{kitai2020designing,baptista2018bayesian} on the Quadratic Unconstrained Binary Optimization (QUBO) matrix $Q$, which seems like a good trade-off for the capacity. A brief introduction to QUBO can be found in section 2. QUBOs are interesting in particular because the optimization $x^*=argmin_x\ E'(x)$ can be done with the help of special computers, for example, a quantum annealer \cite{techreportdwavekelly}.
\ \\
\ \\
Our contribution in this paper is twofold. First, we present the approach ``\textbf{B}lack box \textbf{O}ptimization using the Cross (\textbf{X}) entropy method and \textbf{QUBO}'' (\textit{\textbf{BOX-QUBO}}). It is a black-box optimization algorithm which uses QUBO as the surrogate model and performs significantly better than the state-of-the-art. Second, we present the idea of also solving white-box problems, i.e., problems that could be directly formulated as QUBO, using black-box optimization to reduce the size of the QUBOs to the information-theoretic minimum. Experiments showed that this leads to significantly better solutions for MAX-$k$-SAT.

\section{Background}

In the experiments, we tested our approach \textit{BOX-QUBO} on the domains: Maximum Satisfiability (MAX-SAT), Feedback Vertex Set (FVS) and Maximum Clique (MaxClique). In this chapter, these domains as well as Quadratic Unconstrained Binary Optimization (QUBO) and the Cross-Entropy method are briefly introduced.

\subsection{MAX-SAT}
Satisfiability ($SAT$) is a canonical NP-complete problem \cite{satnpcompletekarp}. Let $V=\{v_1, ..., v_n\}$ be a set of Boolean variables and let $f$ be a Boolean formula represented in conjunctive normal form:
\begin{equation*}
f=\bigwedge_{i=1}^{|f|}{\bigvee_{l \in C_i} l}
\end{equation*}
with $l$ being a literal (a Boolean variable or its negation) and $C_i$ being the i-th clause (a clause is a disjunction of literals). The task is to find an assignment $\underline{V}$ for $V$ with truth values $\{0,1\}$ such that $f(\underline{V})=1$. MAX-SAT is a variant of SAT, where not all clauses have to be satisfied, but only as many as possible. In $k$-SAT, each clause consists of exactly $k$ literals.

\subsection{Feedback Vertex Set (FVS)}
Feedback Vertex Set is an NP-complete problem \cite{satnpcompletekarp}. Let $G=(V, E)$ be a directed graph with vertex set $V$ and edge set $E$ which contains cycles. A cycle is a path (following the edges of $E$) in the graph starting from any vertex $v_S$ such that one ends up back at vertex $v_S$. The path can contain any number of edges. The task in Feedback Vertex Set is to select the smallest subset $V' \subset V$ such that graph $G=(V\backslash V', E')$ is cycle-free, where $E' = \{(v_i, v_j) \in E: v_i \notin V' \land v_j \notin V'\}$.

\subsection{MaxClique}
Maximum Clique is another NP-complete problem \cite{satnpcompletekarp}. Let $G=(V,E)$ be an undirected graph with vertex set $V$ and edge set $E$. A clique in graph $G$ is a subset $V' \subset V$ of the vertices, so all pairs of vertices from $V'$ are connected with an edge. That is, $\forall (v_i,v_j) \in V' \times V': (v_i,v_j) \in E$. The task in MaxClique is to find the largest clique of the graph. The size of a clique is the cardinality of the set $V'$: $|V'|$.

\subsection{Quadratic Unconstrained Binary Optimization (QUBO)}
Let $Q$ be an upper-triangular (n$\times$n)-matrix $Q$ and $x$ be a binary vector of length $n$. The task of Quadratic Unconstrained Binary Optimization \cite{lewis2017quadratic} is to solve the following optimization problem: $x^*=argmin_x (x^TQx)$. QUBO is NP-hard \cite{barahona1982computational} and has been of particular interest recently as special machines, such as the Quantum Annealer \cite{techreportdwavekelly, denchev2016computational} or the Digital Annealer \cite{aramon2019physics}, have been developed to solve these problems. Therefore, in order to solve other problems with these special machines, they must be formulated as QUBO. Numerous problems such as MAX-$k$-SAT \cite{nusslein2022algorithmic,choi2010adiabatic,nusslein2022black, satQAZhengbing,maxksatchancellor,kruger2020quantum}, MaxClique \cite{lucas2014ising,lodewijks2019mapping}, and FVS \cite{lucas2014ising} have already been formulated as QUBO.

\subsection{Cross-Entropy Method}
The cross-entropy method \cite{de2005tutorial} is a general iterative optimization method. The basic idea is, given a parameterized probability distribution $f(x;v)$ and an objective function $g(x)$, to iterate between sampling $X \sim f(x;v)$ from this probability distribution and optimizing this distribution. One starts with random parameters $v$. Then one samples solutions $X \sim f(x;v)$ and sorts them by their values $g(X)$ in increasing order. As a last step, one adjusts the parameters $v$ such that better solutions $x$, which thus had a larger value $g(x)$, get a higher probability $f(x;v)$. This makes it more likely to sample better solutions in the next iteration, starting again with sampling from the probability distribution: $X \sim f(x;v)$, with the new values $v$.

\section{Related Work}

The main Related Work to our approach is Factorization Machine Quantum Annealing (\textit{FMQA}) \cite{kitai2020designing, qakadowaki}. \textit{FMQA}, like our approach, is based on an iteration of three phases: 1) sample from the surrogate model (a QUBO matrix); 2) retrieve the value from the oracle (either via experiments or simulation); 3) update the QUBO matrix. This is in fact a cross-entropy approach with $f$ being the surrogate model (the QUBO matrix). The main difference to our approach is that \textit{FMQA} solely uses regression to train the model, while we use simultaneous regression and classification. The authors subsequently use \textit{FMQA} in their paper to design metamaterials with special properties.
\ \\
\ \\
A similar approach to \textit{FMQA} is Bayesian
Optimization of Combinatorial Structures (BOCS) \cite{baptista2018bayesian} which also uses a quadratic model as a surrogate model. BOCS additionally uses a sparse prior to facilitating optimization. However, BOCS also only uses regression to optimize the model. In the original paper \cite{baptista2018bayesian}, BOCS was solved using only non-quantum methods (including Simulated Annealing \cite{sakirkpatrick}), Koshikawa et al. \cite{koshikawa2021benchmark}, however, also tested BOCS for optimization with a quantum annealer. In \cite{koshikawa2021combinatorial} Koshikawa et al. used BOCS for a vehicle design problem and found that it performed slightly better than a random search.
\ \\
\ \\
In our experiments, we use the white box QUBO formulations for MAX-$k$-SAT \cite{choi2010adiabatic}, FVS \cite{lucas2014ising} and MaxClique \cite{lucas2014ising} as baselines.
\ \\
\ \\
In \cite{huang2021qross}, surrogate QUBO solvers were trained to simplify the optimization of hyperparameters arising in the relaxation of constrained optimization problems. Roch et al. \cite{roch2021cross} used a cross-entropy approach to optimize such hyperparameters.

\section{Black Box Optimization with Cross Entropy and QUBO (BOX-QUBO)}

\textbf{Main idea:} The QUBO matrix is a surrogate model with relatively low capacity since it has only $n(n+1) / 2$ trainable parameters for a matrix of size $(n \times n)$. Previous approaches to black-box optimization with QUBO always attempt full regression on all training data. However, the loss (Mean Squared Error is usually used) becomes larger the more training data is available for the same size of the QUBO matrix due to the limited number of trainable parameters. The main idea behind our approach is to perform a regression only on a small part of the training data and classification on the remaining data since classification requires less capacity of the model than regression.
\ \\
\ \\
For this purpose, the training vectors $x$ are sorted according to their solution quality (also called energy) $E(x)$ and divided into two sets: the set of vectors with higher energies $H$ and the set of vectors with lower energies $L$. There is now a threshold $\tau$ such that:
\begin{equation*}
\begin{split}
    \forall x \in H: E(x) \geq \tau
    \\
    \forall x \in L: E(x) \leq \tau
\end{split}
\end{equation*}
Note that the goal is to find vector $x^*$ with minimum energy: $\forall x: E(x) \geq E(x^*)$. The goal of the classification is to classify whether a vector $x$ is in set $H$ or in set $L$. The regression is performed exclusively on set $L$. The size of $L$ is determined by a hyperparameter $k$. For example, using $k=0.03$, set $L$ contains $3\%$ of all data $D$: $|L|=0.03 \cdot |D|$, thus a much more precise regression is possible. Since $\tau$ is not the hyperparameter but $k$, two vectors $x$ and $x'$ with the same energy ($E(x) = E(x')$) may not be in the same set.
\ \\
\ \\
\textit{BOX-QUBO} requires as input the oracle (black-box function) $E(x)$ and optionally an initial training set $D$. The output will be the vector $x^*$ with minimum energy $y^*=E(x^*)$.
\ \\
\ \\
First, the QUBO matrix $Q$ is randomly initialized (i.e., random values are assigned to all $n(n+1)/2$ parameters of the QUBO matrix). After that, the following three phases alternate:
\begin{enumerate}
    \item Search the $k$ best solution vectors $X^*$ for the current QUBO matrix $Q$ (e.g. using simulated or quantum annealing)
    \item Query the oracle for actual values $E(X^*)$
    \item Append $(X^*, E(X^*))$ to the training set $D$ and retrain $Q$ on $D$
\end{enumerate}
\ \\
The training of $Q$ using the training set $D$ proceeds as follows. First, $D$ is sorted by the energies $E(x)$ and then subdivided into the two non-overlapping sets $H$ and $L$. $H$ contains all vectors $x$ for which $E(x) \geq \tau$ holds. $L$ contains all vectors for which $E(x) \leq \tau$ holds. Details about splitting $H$ and $L$ can be found below.
\ \\
\ \\
After splitting, $Q$ is trained for $nCycles$ iterations, where $nCycles$ is a hyperparameter. In the experiments, we chose values for $nCycles$ between 5 and 10. In each cycle, the \textit{temporary} training set $T$ is created dynamically: all vectors $x \in L$ are always part of $T$, vectors $x \in H$, however, are only part of $T$ if their current prediction $y_{predict}=x^TQx$ is smaller than $\tau$ and thus would violate the classification. The loss function on this temporary training set is now the mean squared error between $y_{predict}=x^TQx$ and $y_{target}$. $y_{target}$, for a vector $x$, is equal to $E(x)$ if $x \in L$ and it is equal to $\tau$ if $x \in H$:
\begin{equation*}
    Q^*=\underset{Q}{\mathrm{argmin}} \left( \mathlarger{\sum}_{x \in T}(y_{predict} - y_{target}) \right) ^2
\end{equation*}
We optimize $Q$ using gradient descent with respect to this loss function. In our experiments, we used a learning rate of 0.005. The other hyperparameter values used in our experiments are listed in Appendix A. The complete algorithm is listed in Algorithm 1.
\ \\
\ \\

\SetKwComment{Comment}{/* }{ */}
\RestyleAlgo{ruled}
\begin{algorithm}[H]
\caption{BOX-QUBO}
\KwData{Set of initial training data $D=\{x,E(x)\}$;  Oracle $E(.)$}
\KwResult{Best solution $(x^*,y^*)$ of the oracle}
\ \\

$Q \gets$ init QUBO matrix\\
$Q \gets train(Q,D)$\\
\ \\

\For{$trainingLength$}{
    $X \gets$ sample best $k$ solutions from $Q$ (e.g. using quantum annealing)\\
    $Y=E(X)$ \tcp*[f]{query oracle $\forall x \in X$}\\
    $D.add(X,Y)$\\
    $Q \gets train(Q,D)$\\
}
\ \\
\KwRet $(x^*,y^*) \in D$ \tcp*[f]{Return best $y$ and corresponding $x$}

\ \\

\DontPrintSemicolon
\SetKwProg{Fn}{Function}{:}{}
\SetKwFunction{FMain}{train}
\Fn{\FMain{$Q$, $D$}}{
    $H, L, \tau \gets$ sort and split $D$ \tcp*[f]{see chapter \textit{Splitting $H$ and $L$}}\\
    \For{$nCycles$}{
        $T, Y_{target} = L, E(L)$ \tcp*[f]{ensure regression $\forall x \in L$}\\
        \For{$x \in H$}{
            $y_{predict}=x^TQx$ \tcp*[f]{ensure classification $\forall x \in H$}\\
            \If{$y_{predict} < \tau$}
            {
                $T \gets x$\\
                $Y_{target} \gets \tau$\\
            }
        }
        $L(x,y_{target})=(x^TQx - y_{target})^2$ \tcp*[f]{the loss function}\\
        optimize $Q$ via gradient descent w.r.t. $\nabla_Q L(T,Y_{target})$
    }
    \KwRet $Q$\;
}

\end{algorithm}
\ \\
\ \\
\ \\
\textbf{Splitting $H$ and $L$:}
\ \\
\ \\
We divide the training data $D$ into the `good' solutions $L$ and the `bad' solutions $H$ based on their energies $E(x)$. For this, we use a hyperparameter $k$ (a percentage): \textit{BOX-QUBO(k\%)}. $k\%$ of the training data are inserted into the set $L$ and the remaining data are inserted into $H$.
\par
In addition, we use the notation \textit{BOX-QUBO(k\%$\backslash$invalids)} to state that all invalid solutions are inserted into the set $H$.
To solve a problem (e.g., MaxClique) with QUBO, the problem must be formulated as QUBO and then the QUBO solution $x$ must be translated back into a problem solution (for example, the set of vertices forming the clique).
We call a QUBO solution $x$ `invalid' if it does not correspond to a valid problem solution (for example, if the QUBO solution $x$ corresponds to a set of vertices $V'$ which is not a clique).

\section{Experiments}

In this paper we have introduced the following two hypotheses:

\begin{enumerate}
\item By using classification and regression simultaneously in our \textit{BOX-QUBO} approach instead of sole regression, lower demands are placed on the capacity of the surrogate model (in our case, the QUBO matrix), which improves the generalization to unknown solutions and thus makes the Black Box Optimization more successful overall (i.e., it finds better solutions). \ \\
\item It can be advantageous to also consider white-box problems as black-box problems to reduce the size of the QUBO matrices to the information-theoretic minimum.
\end{enumerate}
To support these two hypotheses, we tested our approach on four domains: two variants of MAX-$k$-SAT, FVS, and MaxClique. We chose MAX-$k$-SAT because it is a canonical NP-complete problem and the current QUBO formulations (for $k \geq 3$) \cite{nusslein2022algorithmic, lucas2014ising, choi2010adiabatic} do not satisfy the information-theoretic minimum. The information-theoretic minimum for vector $x$ representing the solution for MAX-$k$-SAT with $V$ Boolean variables is the size $|V|$. But the size of $x$ in all QUBO formulations for MAX-$k$-SAT, including the state-of-the-art \cite{nusslein2022algorithmic}, scales linearly in the number of clauses and thus does not correspond to the information-theoretic minimum. Further, we chose FVS because it has invalid QUBO solutions and the QUBO formulation also does not satisfy the information-theoretic minimum. The information-theoretic minimum for FVS corresponds to the cardinality of the vertex set $|V|$. We have chosen MaxClique because there are invalid solutions here as well. But in MaxClique the QUBO formulation \textit{does} correspond to the information-theoretic minimum.

\subsection{Oracles for MAX-$k$-SAT, FVS and MaxClique}
In the experiments, we tested whether the approaches \textit{BOX-QUBO} and \textit{FMQA} are able to solve the domains MAX-$k$-SAT, FVS and MaxClique when these domains are solely accessible as black-box functions. That is, the concrete Boolean formula (for MAX-$k$-SAT) or the concrete graph (for FVS and MaxClique) are unknown.

The oracles return an indirectly proportional value to the solution quality and the value 1 if the QUBO solution does not translate into a valid problem solution. The goal of \textit{BOX-QUBO} and \textit{FMQA} was then to minimize the black-box functions (oracles). We have chosen as oracles:
\ \\

$SAT_{oracle}(x) = -(\#\ satisfied\ clauses\ with\ variable\ assignment\ x)$
\ \\

$FVS_{oracle}(x) = if\ G \backslash x\ still\ has\ cycles\ then\ 1\ else\ (\Sigma_i x_i - |V|)$
\ \\

$MaxClique_{oracle}(x) = if\ x\ is\ a\ valid\ clique\ then\ (-\Sigma_i x_i)\ else\ 1$
\ \\
\ \\
$G$ is the graph and $|V|$ is the number of vertices in the graph. $(\Sigma_i x_i)$ calculates the number of 1s in the vector $x$, which represents in MaxClique the size of the clique and in FVS the number of deleted vertices.

\subsection{Results for MAX-$k$-SAT, FVS and MaxClique}

We tested \textit{BOX-QUBO} and \textit{FMQA} on a total of four domains:
\begin{enumerate}
    \item Max-4-SAT with $(V=20, C=300)$
    \item Max-4-SAT with $(V=30, C=400)$
    \item FVS with $(V=25, E=200)$
    \item MaxClique with $(V=30, E=350)$
\end{enumerate}
\ \\
In MAX-$k$-SAT, $V$ is the number of boolean variables and $C$ is the number of clauses. In FVS and MaxClique, $V$ is the number of vertices of the graph and $E$ is the number of edges. The values (for example, the number of variables and clauses for Max-4-SAT) were chosen so that the overall task was difficult. For example, if $(V=20, C=50)$ had been chosen for Max-4-SAT, problem instances of this type would be easy to solve as there are numerous variable assignments for $V$ that satisfy all $50$ clauses.
\ \\
\ \\
For each of the four domains, $15$ random instances (with different seeds) were created. That is, for Max-4-SAT $15$ random formulas were created and for FVS and MaxClique $15$ different graphs were generated each. Then, the \textit{BOX-QUBO} and \textit{FMQA} approaches were applied to each of the $15$ instances (i.e. on the corresponding oracle functions).
\ \\
\ \\
We first generated between 2,000 (for domain 1) and 10,000 (for domains 2, 3 and 4) random QUBO solutions as the initial training set. Both algorithms used the same initial training set. The best solution from this initial set served as the baseline $base$. In addition, specialized white-box methods were used to determine the true optimal solution to each instance, which we refer to as $optimum$. To better compare performance between domains, we normalized the best solution found for each of the \textit{BOX-QUBO} and \textit{FMQA} approaches, using the formula: $f(y^*)=(y^*-base)/(optimum-base)$. This gives $f(y^*)=1$ if $y^*=optimum$ and $f(y^*)=0$ if $y^*=base$, where $y^*$ is the value of the best solution ($x^*$) found so far. We ran the black-box optimization for $trainingLength$ iterations. One iteration consists of sampling the current best solutions and retraining the model until the loss converged. The results are shown in \textbf{Figure 1}.

\begin{figure}[H]
\centering
\includegraphics[scale=0.45]{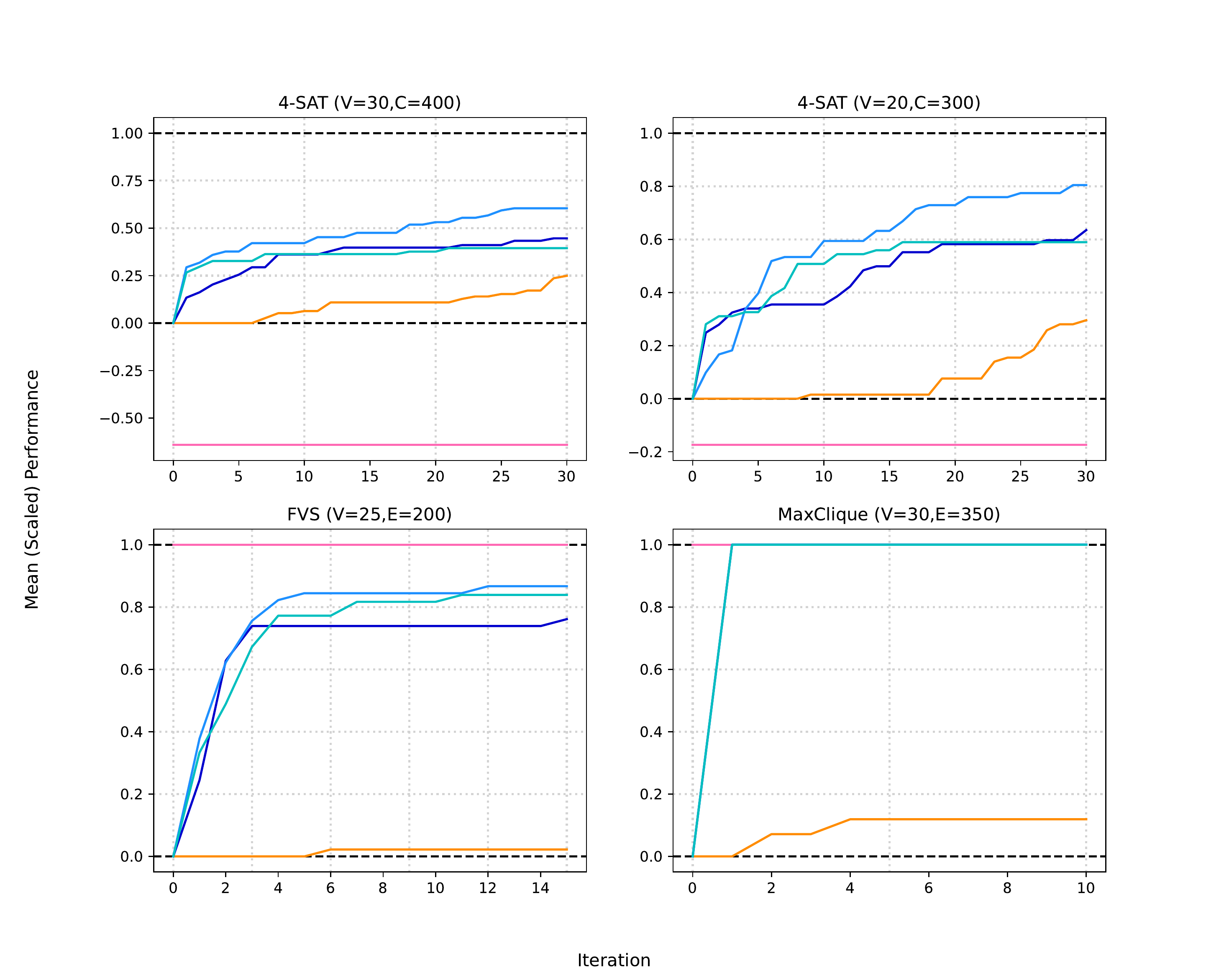}
\includegraphics[scale=0.62]{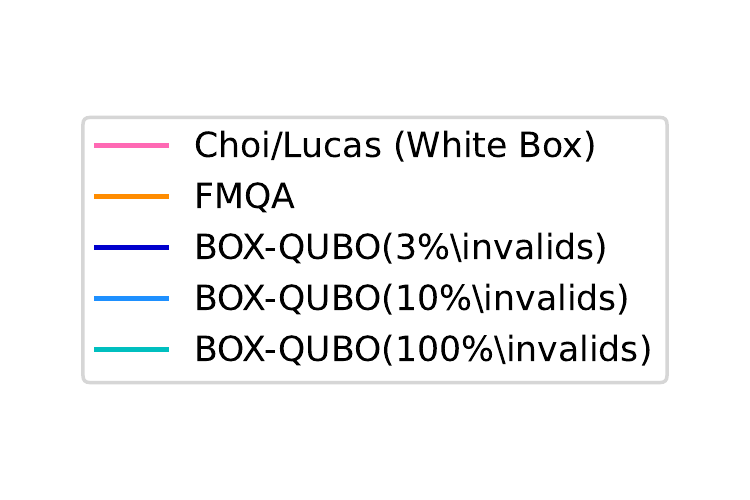}
\caption{Normalized performance of \textit{BOX-QUBO}, \textit{FMQA} \cite{kitai2020designing} and the white-box solutions of Choi \cite{choi2010adiabatic} and Lucas \cite{lucas2014ising}. The x-axes describe the iteration number and the y-axes describe the mean performance over 15 instances, where the quality (oracle value) $y^*$ of the best-found solution $x^*$ is normalized: $f(y^*)$ such that $f(y^*)=1$ if $y^*=optimum$ and $f(y^*)=0$ if $y^*=base$. In MaxClique, the graphs of \textit{BOX-QUBO(100\%$\backslash$invalids)}, \textit{BOX-QUBO(10\%$\backslash$invalids)} and \textit{BOX-QUBO(3\%$\backslash$invalids)} overlap.}
\end{figure}
\ \\
The x-axes of the diagrams represent the iteration number and the y-axes represent the average value of $f(y^*)$ over the 15 instances: $1/15 \sum_{i=1}^{15}f(y^*_i)$, where $y^*$ is the value of the best solution found for problem instance $i$ in the current iteration, making all graphs monotonically increasing.
\ \\
\ \\
In addition to \textit{BOX-QUBO} and \textit{FMQA}, the respective instances were also solved as white-box problems for comparison. For Max-4-SAT this was the QUBO formulation of Choi \cite{choi2010adiabatic}, for FVS and MaxClique these were the QUBO formulations of Lucas \cite{lucas2014ising}. For \textit{FMQA} we used the GitHub implementation: [\url{https://github.com/tsudalab/fmqa}]. For sampling from the QUBOs we used simulated annealing. Both approaches (\textit{BOX-QUBO} and \textit{FMQA}) were trained in each iteration until the loss converged.
\ \\
\ \\
There are two things worth mentioning in the results. First, if we look at the two Max-4-$SAT$ results, we see that the black-box optimization (both \textit{BOX-QUBO} and \textit{FMQA}) found significantly better solutions than Choi's white-box solution. Using the black-box optimization, we have reduced the size of the QUBO matrix to the information-theoretic minimum. For example, in the first experiment, we considered formulas with $V=30$ variables and $C=400$ clauses. Here, Choi's QUBO matrix had size $n=4 \cdot 400=1600$, while the QUBO matrices for \textit{BOX-QUBO} and \textit{FMQA} corresponded to the information-theoretic minimum ($n=V=30$). However, solving with black-box methods is not always better than solving with white-box methods, as can be seen in the results for \textit{FVS}.
\ \\
\ \\
The second conclusion that can be drawn from our experiments is that \textit{BOX-QUBO} was always more successful than \textit{FMQA}. In \textit{FVS}, for example, \textit{BOX-QUBO(10\%$\backslash$invalids)} reached a mean performance of roughly $0.87$ after 15 iterations while \textit{FMQA} only reached roughly $0.03$. A similar picture emerged for MaxClique, with all \textit{BOX-QUBO} variants reaching the optimum already after the first iteration. The code for \textit{BOX-QUBO} is available on GitHub
[\url{https://github.com/JonasNuesslein/BOX-QUBO}]

\subsection{Dimensionality Reduction}
By solving problems using black-box optimization, the size of QUBOs can be reduced to the information-theoretic minimum.  \textbf{Table 1} lists the dimensionality reduction of the QUBO matrices when white-box problems from MAX-$k$-SAT, FVS or MaxClique are solved using black-box optimization instead. Max-4-$SAT$ with $(V=20,C=300)$ reduces from $n=1200$ (the state-of-the-art QUBO formulation for 4-$SAT$ \cite{choi2010adiabatic}) to $n=20$ and Max-4-$SAT$ with $(V=30,C=400)$ reduces from $n=1600$ to $n=30$. $FVS$ with $(V=25,E=200)$ reduces from $n=25 \cdot (25+1)=650$ to $n=25$. The white-box QUBO formulation for MaxClique already encodes the problems using the information-theoretic minimum.
\begin{table}[H]
\begin{center}
\begin{tabular}{ |c|c|c|c|c| } 
 \hline
    & Max-4-$SAT$ & Max-4-$SAT$ & $FVS$ & $MaxClique$ \\
    & $(30,400)$ & $(20,300)$ & $(25,200)$ & $(30,350)$ \\ \hline
    Black Box & $n=30$ & $n=20$ & $n=25$ & $n=30$ \\
    (\textit{FMQA} or \textit{BOX-QUBO}) & $\mathcal{O}(V)$ & $\mathcal{O}(V)$ & $\mathcal{O}(V)$ & $\mathcal{O}(V)$ \\ \hline
    White Box & $n=1600$ & $n=1200$ & $n=650$ & $n=30$ \\
    (Choi/Lucas) & $\mathcal{O}(kC)$ & $\mathcal{O}(kC)$ & $\mathcal{O}(V^2)$ & $\mathcal{O}(V)$ \\ \hline
\end{tabular}
\newline
\newline
\end{center}
\caption{\label{tab:tab1}Listed are the sizes $n$ of the $(n \times n)$ QUBO matrices and the scaling of the problems in $\mathcal{O}$-notation for the four domains for white-box or black-box, respectively. The white-box QUBO formulations for Max-4-$SAT$ and $FVS$, unlike $MaxClique$, do not encode the problems using the information-theoretic minimum.}
\end{table}

\section{Conclusion and Future Work}

In this paper, we studied black-box optimization using QUBOs as surrogate models. We presented the \textit{BOX-QUBO} approach, which is characterized by the idea to use a simultaneous classification and regression, instead of regression alone. The reason for this is that classification places fewer demands on the capacity of the model (i.e., the QUBO matrix), which has a positive effect on the accuracy of the remaining regression and thus on generalization, leading to better solutions overall.
\ \\
\ \\
We have tested our approach on the MAX-$k$-SAT, $FVS$, and $MaxClique$ domains, and the experiments showed that \textit{BOX-QUBO} consistently outperformed \textit{FMQA}. There are white-box QUBO formulations for these domains,  but we think the superiority of \textit{BOX-QUBO} will also hold for problems for which no white-box QUBO formulations exist. However, important future work is to apply \textit{BOX-QUBO} to (real world) black-box functions for which no white-box QUBO formulations exist.
\ \\
\ \\
Besides introducing \textit{BOX-QUBO}, we also presented the idea of solving white-box problems using black-box optimization to reduce the size of the QUBO matrices to the information-theoretic minimum. The experiments on MAX-4-$SAT$ showed that the solutions found using black-box optimization were significantly better than those found using the white-box QUBO formulation.
\ \\
\ \\
For future work, we propose to introduce ancilla QUBO variables to dynamically control the capacity of the surrogate model.
\ \\
\ \\
\textbf{Acknowledgement}
\ \\
\ \\
This publication was created as part of the Q-Grid project (13N16179) under the ``quantum technologies - from basic research to market'' funding program, supported by the German Federal Ministry of Education and Research.

%
%
%
\newpage
\bibliographystyle{splncs04}
\bibliography{references}

\newpage

\appendix

\section{Hyperparameters}

Here we list the hyperparameters for \textit{BOX-QUBO} for the four domains. In parentheses are the values for the one-time initial training.
\ \\
\ \\
\textbf{Max-4-SAT (V=20,C=300)}:
\begin{itemize}
    \item $nCycles=5\ (30)$
    \item $nEpochs=150\ (600)$
    \item $init\_training\_size = 2,000$
\end{itemize}
\ \\
\textbf{Max-4-SAT (V=30,C=400)}:
\begin{itemize}
    \item $nCycles=5\ (30)$
    \item $nEpochs=150\ (600)$
    \item $init\_training\_size = 10,000$
\end{itemize}
\ \\
\textbf{FVS (V=25,C=200)}:
\begin{itemize}
    \item $nCycles=10\ (30)$
    \item $nEpochs=150\ (400)$
    \item $init\_training\_size = 10,000$
\end{itemize}
\ \\
\textbf{MaxClique (V=30,C=350)}:
\begin{itemize}
    \item $nCycles=10\ (30)$
    \item $nEpochs=150\ (400)$
    \item $init\_training\_size = 10,000$
\end{itemize}
\ \\
For \textit{FMQA}, the implementation of [\url{https://github.com/tsudalab/fmqa}] was used. For MAX-$k$-SAT PySAT was used to determine the optima and for FVS [\url{https://github.com/yanggangthu/FVS_python}] was used.

\end{document}